\documentclass[letterpaper]{article}

\usepackage[utf8]{inputenc} 
\usepackage[T1]{fontenc}    
\usepackage{hyperref}       
\usepackage{url}            
\usepackage{booktabs}       
\usepackage{amsfonts}       
\usepackage{nicefrac}       
\usepackage{microtype}      
\usepackage{xcolor}         
\usepackage{graphicx}       
\usepackage{subcaption}     
\usepackage{tikz}
\usetikzlibrary{positioning, arrows.meta, backgrounds, decorations.pathreplacing}                                  
\usetikzlibrary{positioning}

\usepackage{natbib}  
\usepackage{alifeconf}

\usepackage{amsmath}        
\usepackage{amssymb}        
\usepackage{amsthm}
\usepackage{mathtools}      
\usepackage{bm}             

\usepackage{algorithm}
\usepackage{algpseudocode}

\usepackage{cleveref}       
\usepackage{enumitem}       
\usepackage[bottom]{footmisc}

\newcommand{\world}{\Omega}

\newcommand\blfootnote[1]{%
  \begingroup
  \renewcommand\thefootnote{}\footnote{#1}%
  \addtocounter{footnote}{-1}%
  \endgroup
}

\usepackage{titlesec}
\titlespacing*{\paragraph}{0pt}{0.4em}{0.7em}

\title{Evolving Many Worlds:\\ Towards Open-Ended Discovery in Petri Dish NCA via Population-Based Training}

\author{
    Uljad Berdica$^{1}$$^{*}$ \quad Jakob Foerster$^1$ \quad Frank Hutter$^{4,2,3}$ \quad Arber Zela$^{5}$ $^{\dagger, *}$ \\
    \mbox{}\\
    $^1$FLAIR, University of Oxford \quad $^2$ELLIS Institute Tübingen \quad $^3$University of Freiburg \quad $^4$Prior Labs \quad $^5$EPFL \\
    uljadb@robots.ox.ac.uk \quad\quad arber.zela@epfl.ch
} 

\begin{document}
\maketitle

\begin{abstract}
The generation of sustained, open-ended complexity from local interactions remains a fundamental challenge in artificial life. Differentiable multi-agent systems, such as Petri Dish Neural Cellular Automata (PD-NCA), exhibit rich self-organization driven purely by spatial competition; however, they are highly sensitive to hyperparameters and frequently collapse into uninteresting patterns and dynamics, such as frozen equilibria or structureless noise. In this paper, we introduce PBT-NCA, a meta-evolutionary algorithm that evolves a population of PD-NCAs subject to a composite objective that rewards both historical behavioral novelty and contemporary visual diversity. Driven by this continuous evolutionary pressure, PBT-NCA spontaneously generates a plethora of emergent lifelike phenomena over extended horizons—a hallmark of true open-endedness. Strikingly, the substrate autonomously discovers diverse morphological survival and self-organization strategies. We observe highly regular, coordinated periodic waves; spore-like scattering where homogeneous groups eject cell-like clusters to colonize distant territories; and fluid, shape-shifting macro-structures that migrate across the substrate, maintaining stable outer boundaries that enclose highly active interiors. By actively penalizing monocultures and dead states, PBT-NCA sustains a state of effective complexity that is neither globally ordered nor globally random, operating persistently at the ``edge of chaos''.
\end{abstract}


Data/Code available at: \href{https://github.com/arberzela/pbt-nca}{\texttt{arberzela/pbt-nca}}
\blfootnote{$^*$Equal Contribution. $^\dagger$ Work done while at ELLIS Institute Tübingen.}
\blfootnote{\textcopyright  2026 Berdica, Foerster, Hutter, Zela. Published under a Creative Commons Attribution 4.0 International (CC BY 4.0) license.}

\begin{figure}[t!]
\centering
\resizebox{0.99\linewidth}{!}{
\begin{tikzpicture}[
    world/.style={draw, thick, minimum size=0.9cm, rounded corners=2pt},
    elite/.style={world, fill=green!25, draw=green!60!black},
    mid/.style={world, fill=yellow!20, draw=yellow!60!black},
    bottom/.style={world, fill=red!20, draw=red!60!black},
    dead/.style={world, fill=gray!15, draw=gray!50, dashed},
    newworld/.style={world, fill=blue!15, draw=blue!60!black},
    arrow/.style={->, >=stealth, thick},
    copyarrow/.style={arrow, green!60!black},
    mutarrow/.style={arrow, blue!60!black, dashed},
    label/.style={font=\scriptsize\sffamily},
    phase/.style={font=\small\sffamily\bfseries, text=black},
    annot/.style={font=\tiny\sffamily, text=gray!30!black},
]

\node[phase, font=\Large] at (2.4, 3.2) {\textbf{\texttt{Meta-iteration} $\mathtt{t}$}};

\node[phase, text=black!70, font=\footnotesize] at (-1.8, 2.3) {\textbf{1. Rollout \&}};
\node[phase, text=black!70, font=\footnotesize] at (-1.9, 2.0) {\textbf{Score}};

\node[elite, label=below:{\scriptsize $W_1$}] (w1) at (0, 2.2) {\includegraphics[width=0.9cm,height=0.9cm]{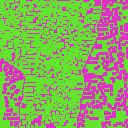}};
\node[elite, label=below:{\scriptsize $W_2$}] (w2) at (1.2, 2.2) {\includegraphics[width=0.9cm,height=0.9cm]{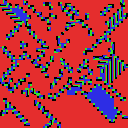}};
\node[mid,   label=below:{\scriptsize $W_3$}] (w3) at (2.4, 2.2) {\includegraphics[width=0.9cm,height=0.9cm]{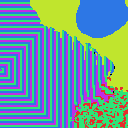}};
\node[bottom,label=below:{\scriptsize $W_4$}] (w4) at (3.6, 2.2) {\includegraphics[width=0.9cm,height=0.9cm]{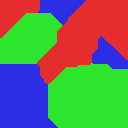}};
\node[bottom,label=below:{\scriptsize $W_5$}] (w5) at (4.8, 2.2) {\includegraphics[width=0.9cm,height=0.9cm]{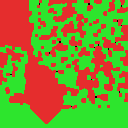}};

\draw[arrow, gray] (0, 1.64) -- (0, -0.5);
\draw[arrow, gray] (1.2, 1.64) -- (1.2, -0.5);
\draw[arrow, gray] (2.4, 1.64) -- (2.4, -0.5);

\node[phase, text=black!70, font=\footnotesize] at (-1.8, 0.72) {\textbf{2. Archive}};
\node[phase, text=black!70, font=\footnotesize] at (-1.65, 0.42) {\textbf{Update}};
\draw[thick, rounded corners=3pt, fill=green!8, draw=green!40!black]
    (-0.5, 0.3) rectangle (5.3, 0.9);
\node[annot, font=\small] at (2.4, 0.6) {$\mathcal{A} \leftarrow \mathcal{A} \cup \{d_1, d_2\}$ (top-$m$ worlds, FIFO)};

\node[phase, text=black!70, font=\footnotesize] at (-1.8, -1.00) {\textbf{3. Exploit-}};
\node[phase, text=black!70, font=\footnotesize] at (-1.65, -1.30) {\textbf{Explore}};

\node[elite] (k1) at (0, -1.1) {\includegraphics[width=0.9cm,height=0.9cm]{figures/frames/pbt-nca-n7/output_340.png}};
\node[elite] (k2) at (1.2, -1.1) {\includegraphics[width=0.9cm,height=0.9cm]{figures/frames/pbt-nca-largerspace/output_086.png}};
\node[mid]   (k3) at (2.4, -1.1) {\includegraphics[width=0.9cm,height=0.9cm]{figures/frames/bad_frames/output_141.png}};
\node[annot] at (1.2, -1.85) {};

\node[dead] (d4) at (3.6, -1.1) {};
\node[dead] (d5) at (4.8, -1.1) {};
\node[annot] at (4.2, -1.85) {};

\draw[copyarrow, bend right=25] (k1.south) to
    node[annot, below, sloped] {} (d4.south);
\draw[copyarrow, bend right=20] (k2.south) to
    node[annot, below, sloped] {} (d5.south);

\node[phase, text=black!70] at (2.4, -2.6) {Create Offsprings};

\node[elite, minimum size=0.7cm] (src) at (-1.0, -3.4) {};
\node[annot, below=1pt of src] {};

\draw[copyarrow] (src) -- node[annot, above] {Copy} (0.33, -3.4);

\node[world, fill=green!10, draw=green!40!black, minimum size=0.7cm]
(copied) at (0.7, -3.4) {};

\draw[arrow, orange!70!black] (copied) -- node[annot, above]
{Crossover} (2.12, -3.4);

\node[world, fill=orange!10, draw=orange!60!black, minimum size=0.7cm]
(crossed) at (2.5, -3.4) {};

\draw[mutarrow] (crossed) -- node[annot, above]
{Mutate} (3.8, -3.4);

\node[world, fill=blue!10, draw=blue!50!black, minimum size=0.7cm]
(mutated) at (4.2, -3.4) {};

\draw[arrow, purple!70!black, densely dotted] (mutated) -- node[annot, above]
{Perturb} (5.43, -3.4);

\node[newworld, minimum size=0.7cm] (child) at (5.8, -3.4) {};
\node[annot, below=1pt of child] {};

\node[phase, font=\Large] at (2.4, -4.4) {\textbf{\texttt{Meta-iteration $\mathtt{t{+}1}$}}};

\node[elite,    label=below:{\scriptsize $W_1$}]  at (0, -5.4) {\includegraphics[width=0.9cm,height=0.9cm]{figures/frames/pbt-nca-n7/output_340.png}};
\node[elite,    label=below:{\scriptsize $W_2$}]  at (1.2, -5.4) {\includegraphics[width=0.9cm,height=0.9cm]{figures/frames/pbt-nca-largerspace/output_086.png}};
\node[mid,      label=below:{\scriptsize $W_3$}]  at (2.4, -5.4) {\includegraphics[width=0.9cm,height=0.9cm]{figures/frames/bad_frames/output_141.png}};
\node[newworld, label=below:{\scriptsize $W_4'$}] at (3.6, -5.4) {\includegraphics[width=0.9cm,height=0.9cm]{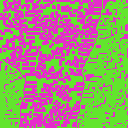}};
\node[newworld, label=below:{\scriptsize $W_5'$}] at (4.8, -5.4) {\includegraphics[width=0.9cm,height=0.9cm]{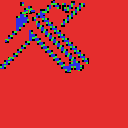}};


\begin{scope}[shift={(6.4, 2.63)}]
\node[draw, rounded corners, fill=white, inner sep=2.9pt] (legendbox) at (0.6, -2.0) {
\begin{tikzpicture}

\node[phase] at (0.7, 0.6) {Legend};

\node[elite, minimum size=0.5cm] at (0, 0) {};
\node[annot, right] at (0.4, 0) {Elite (top)};

\node[mid, minimum size=0.5cm] at (0, -0.55) {};
\node[annot, right] at (0.4, -0.55) {Mid-rank};

\node[bottom, minimum size=0.5cm] at (0, -1.1) {};
\node[annot, right] at (0.4, -1.1) {Low-rank};

\node[newworld, minimum size=0.5cm] at (0, -1.65) {};
\node[annot, right] at (0.4, -1.65) {New child};

\node[dead, minimum size=0.5cm] at (0, -2.2) {};
\node[annot, right] at (0.4, -2.2) {Discarded};

\end{tikzpicture}
};
\end{scope}

\end{tikzpicture}
}
\caption{%
Overview of one PBT-NCA meta-iteration. Each colored square represents a
PD-NCA world (weights, optimizer, world state, and hyperparameters). In
\textbf{Stage~1}, all worlds are rolled out and scored by a fitness combining
archive-based novelty and DINOv2 cross-world temporal diversity. In \textbf{Stage~2}, the top-$m$ behavior descriptors are added to a FIFO archive. In \textbf{Stage~3}, the lowest-fitness worlds are replaced by mutated copies of elite parents.}
\label{fig:pbt_overview}
\vspace{-10pt}
\end{figure}

\begin{figure*}[t]
    \centering
    \includegraphics[width=0.90\linewidth]{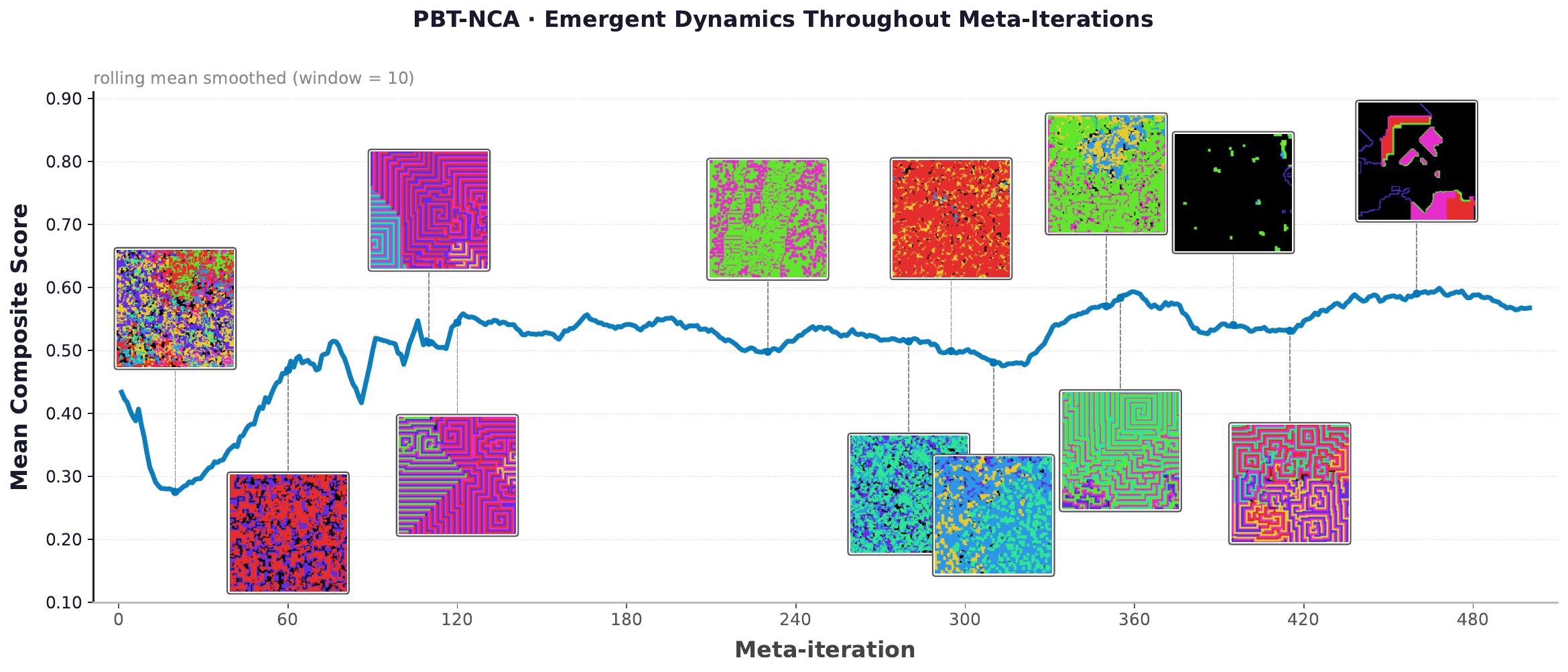}
    \caption{PBT-NCA evolving a population of 30 PD-NCA worlds, each with 7 NCA agents competing for territory. We plot the composite score function over meta-iterations and rollouts from the highest scoring world. Novel dynamics emerge from agentic competition in an open-ended progression. Videos available at: \href{https://pbt-nca.github.io}{\texttt{pbt-nca.github.io}}.}
    \label{fig:frames_fitness_7}
    \vspace{-1mm}
\end{figure*}

\section{Introduction}

The quest to understand how simple, local interactions give rise to sustained, open-ended complexity is a central challenge unifying evolutionary biology, artificial life, and non-equilibrium physics~\citep{wolfram83,wolfram1984cellular,holland92,taylor2016open}. In nature, open-endedness is observed as the endless production of novel and non-trivial structures, allowing for the accumulation of behavioral and structural diversity~\citep{Bedau1996MeasurementOE,taylor2016open,lehman2011novelty,stanley2019oe,stanley2019openendedness}. Replicating this phenomenon for artificial systems entails a fundamental transition from static optimization to perpetual morphological and functional innovation, a property believed to be essential for achieving artificial superhuman intelligence (ASI)~\citep{hughes2024open}. Many works suggest that this requires operating at the ``edge of chaos,’’ a delicate phase transition where dynamics are neither globally ordered and highly compressible, nor random and fully structureless~\citep{langton1990, kauffman1993origins,zhang2025intelligence}. 

Cellular Automata (CA)~\citep{wolfram83,wolfram1984cellular,Wolfram2002} have long served as a powerful framework for modeling emergent phenomena from simple, local interactions. Building on this discrete foundation, continuous-state CA like Lenia~\citep{Chan2018LeniaB, chan2020expanded, scaling_lenia_23,flowlenia} demonstrated that hand-designed mathematical rules in continuous domains could produce a broad spectrum of lifelike patterns. More recently, this concept has been extended to deep learning through Neural Cellular Automata (NCA)~\citep{mordvintsev2020growing}. In NCAs, the local update rule is learned by training a neural network to grow, maintain, and regenerate target morphologies on a 2D~\citep{niklasson2021self-organising,randazzo2020selfclassifying,randazzo2021adversarial} or 3D~\citep{sudhakaran2021growing} grid.

\textit{Petri Dish Neural Cellular Automata} (PD-NCA)~\citep{pdnca2025} extend conventional NCAs to a competitive, continual multi-agent setting, in which multiple NCA agents compete for territorial expansion within a shared substrate, hereafter referred to as a \textit{world}. However, in the absence of carefully designed loss functions or well-tuned hyperparameters, the high-dimensional parameter space of PD-NCAs contains a lot of regions that lead to uninteresting, non-complex behaviors where system dynamics often freeze into static equilibria, devolve into high-entropy noise, or collapse to monocultures in which a single NCA agent uniformly dominates the substrate by occupying all the grid cells. 

To overcome these limitations, we introduce \textit{Population-Based Training of Petri Dish Neural Cellular Automata} (\textbf{PBT-NCA}). As illustrated in~\autoref{fig:pbt_overview}, PBT-NCA applies population-level exploit-and-explore cycles~\citep{jaderberg2017pbt} to evolve a parallel population of PD-NCA worlds under a novelty-driven selection pressure. By combining an archive-based score function~\citep{lehman2011novelty} computed from handcrafted descriptors with a population-level visual diversity score derived from a frozen DINOv2 encoder~\citep{oquab2023dinov2}, we score each world by how different it is from both the history of past behaviors and the rest of the current population, rather than rewarding them for reaching a fixed target~\citep{mordvintsev2020growing,niklasson2021self-organising,randazzo2021adversarial}. While previous methods like ASAL~\citep{kumar2025automating} or ASAL$++$~\citep{baid2025guiding} use foundation models for outer-loop search over frozen world configurations, PBT-NCA directly applies selection pressure to a population where individual agents \emph{interact} with each-other and \textit{adapt} their local update rules.

\paragraph{Emergent Symbiosis at the Edge of Chaos.}
Under our novelty-driven selection mechanism, PBT-NCA evolves worlds that spontaneously generate strikingly autopoietic lifelike entities (Figure~\ref{fig:frames_fitness_7}), where structural regularities persist in the substrate, rather than collapsing to trivial equilibria or noise. We observe emergent phenomena including fluid, shape-shifting macro-structures with coordinated motion, distributed territorial clusters self-replicating via long-range colonization, and the gradual formation of spatial organization. We further identify decentralized, trail-like locomotion patterns arising purely from local interactions, where agents start coordinating without centralized control (Figure~\ref{fig:pd-nca-ants}).
Crucially, these behaviors emerge without explicit objectives for cooperation or structure, but as a byproduct of continual pressure for novelty and adaptation. By evolving a population of worlds, PBT-NCA acts as an engine of open-endedness, enabling the ongoing accumulation of behavioral and structural novelty over long time horizons. The resulting worlds reside at the ``edge of chaos'', a regime between order and randomness where rich, adaptive dynamics emerge~\citep{langton1990,kauffman1993origins,Mitchell1994Revisiting}.
\section{Population-Based Training of Petri Dish NCAs}
\label{sec:pbt-nca}

We extend the PD-NCA substrate \citep{pdnca2025} with a population-based meta-optimization loop designed to favor continual discovery of new ecological regimes~\citep{jaderberg2017pbt,liu2019emergent,leniabreeder,scaling_lenia_23}. The main motivation is that in single-world training, once a particular interaction pattern becomes dominant, gradient descent tends to reinforce it, often leading to monocultures, frozen equilibria, or unstructured fluctuations. We therefore make selection explicitly non-stationary. Rather than rewarding one fixed target behavior, we score each world relative to both a memory of past behaviors and the diversity of the current population~\citep{lehman2011novelty,pugh2016qdmaps}. The goal is thus not to converge to one optimum, but to continually generate and maintain a population of worlds that remain behaviorally novel, visually distinct, and dynamically active. We name our method PBT-NCA and provide the pseudocode in Algorithm~\ref{alg:pbt-nca}.

\subsection{The PD-NCA Substrate}

A PD-NCA~\footnote{\url{https://pub.sakana.ai/pdnca/}} world $\world$ is a differentiable multi-agent cellular system where $N$ neural networks (agents) act as distinct ``species'' competing for territory on a shared grid. 
At time $t$, the global state $X^t \in \mathbb{R}^{H \times W \times C}$, where $H$ and $W$ denote the grid height and width, consists of a feature vector $x_{u,v}^t \in \mathbb{R}^C$ at each spatial position $(u,v)$. This vector is explicitly partitioned into attack channels $\mathbf{a}_{u,v}^t \in \mathbb{R}^{C_a}$, defense channels $\mathbf{d}_{u,v}^t \in \mathbb{R}^{C_d}$, and a hidden state $\mathbf{h}_{u,v}^t \in \mathbb{R}^{C_h}$, where $C = C_a + C_d + C_h$, that allows agents to process information separately: $x_{u,v}^t = [\mathbf{a}_{u,v}^t; \mathbf{d}_{u,v}^t; \mathbf{h}_{u,v}^t]$. 
Alongside these cell-level traits, the system maintains a global ``aliveness'' mask $A^t \in \mathbb{R}^{N \times H \times W}$ that tracks exactly which agents occupy which cells.
At every time step, each living agent $k$ observes its immediate Moore neighborhood $\mathcal{N}_{u,v}(X^t)$ centered at $(u,v)$. Based on this local view, its internal neural network parameterized by $\theta_k$ decides how to act, proposing updates $\Delta x_{u,v}^{t,k} = f_{\theta_k}\!\left(\mathcal{N}_{u,v}(X^t)\right)$.
To ensure the agents never stop adapting, a \textit{background environment} ($k=0$) akin to a harsh climate or natural decay, constantly proposes its own random, static updates via a $L_2$-normalized uniform noise tensor $\Delta x_{u,v}^{t,0} = E_{u,v}$, drawn from $\mathcal{U}(-1, 1)$, forcing agents to maintain active defenses just to survive and effectively preventing evolutionary stagnation.

Multiple agents can allocate resources, e.g. for attack or defense, to the same cell; the environment can also update the cell's conditions at the same time. This competition between agents $k$ and $l$ for cell $(u, v)$ is formalized via pairwise cosine similarities: $\phi_{kl}(u,v) = \langle \mathbf{a}_{u,v}^{t,k}, \mathbf{d}_{u,v}^{t,l} \rangle - \langle \mathbf{d}_{u,v}^{t,k}, \mathbf{a}_{u,v}^{t,l} \rangle$. The total competitive strength of agent $k$ at cell $(u,v)$ is the sum of its pairwise interactions with every other agent and with the environment: $\Psi_k(u,v) = \sum_{l \neq k} \phi_{kl}(u,v) + \phi_{\text{k,env}}(u,v)$. Crucially, this competition is not strictly winner-takes-all; strengths are normalized into contribution weights $w_{u,v}^{t,k} = softmax( \Psi_k(u,v) )$, meaning the cell's new state is a weighted blend of the competing proposals: $x_{u,v}^{t+1} = \operatorname{clip}\!\left( x_{u,v}^{t} + \sum_{k=0}^{N} w_{u,v}^{t,k}\,\Delta x_{u,v}^{t,k} \right)$, with clipping enforcing valid ranges. These shares also dictate an agent's ongoing aliveness, updated via $A_{u,v}^{t+1,k} = w_{u,v}^{t,k}$ if $w_{u,v}^{t,k} > \alpha$, and $0$ otherwise. 
If an agent's share drops below $\alpha = 0.4$, it dies off in that location and its remaining influence is redistributed. This specific threshold allows up to two different agents to coexist in a single cell simultaneously, ensuring that territorial boundaries remain fluid and dynamic.

A PD-NCA world evolves on two coupled levels: the grid state changes through local interactions, while the agents concurrently adapt their internal parameters. 
Every $\tau$ rollout steps, each agent $k$ seeks territorial expansion by maximizing its log-aliveness across the entire grid: $\mathcal{L}_k = - \log \left( \sum_{u,v} A_{u,v}^{t+\tau, k} \right)$, and updating their parameters via gradient descent, $\theta_k \leftarrow \theta_k - \eta_k \nabla_{\theta_k}\mathcal{L}_k$, where $\eta_k$ is the current learning rate. This simple, gradient-based drive for growth naturally forces the substrate to self-organize into complex dynamics.

\begin{algorithm}[t]
\caption{PBT-NCA pseudocode}
\label{alg:pbt-nca}
\begin{algorithmic}[1]
\Require Population size $P$, meta-iterations $T$, PD-NCA train epochs $T_{world}$, exploit interval $K$, replacement fraction $\rho$, archive increment $m$, crossover and perturb probabilities $P_{cross}$, $P_{pert}$
\State Init. random population $\{\world_i\}_{i=1}^{P}$, archive $\mathcal{A} \gets \varnothing$
\For{$t = 1$ \textbf{to} $T$}
    \State Train each $\Omega_i$ for $T_{world}$ steps; compute score $\mathcal{F}_i$
    \State Add top-$m$ descriptors to $\mathcal{A}$
    \If{$t \bmod K = 0$}
        \ForAll{$\Omega_c$ in bottom $\rho$ fraction}
            \State Sample parent $\Omega_p$ from top $\rho$ fraction
            \State Copy parent to child: $\Omega_c \gets \Omega_p$
            \State Crossover each hparam w.p.\ $P_{cross}$ 
            \State Perturb each hparam by $\times(1\pm0.2)$ w.p.\ $P_{pert}$
            \State Add Gaussian noise to $\Omega_c$ weights
        \EndFor
    \EndIf
\EndFor
\end{algorithmic}
\end{algorithm}

%
%
%

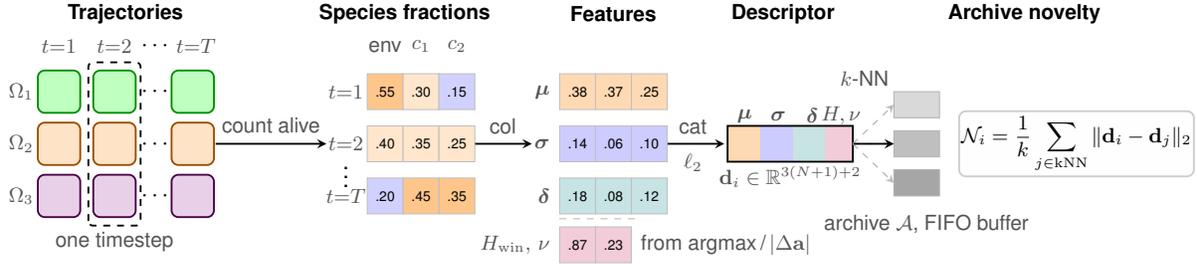
\begin{figure*}[t!]
\centering
\resizebox{0.9\linewidth}{!}{%

\begin{tikzpicture}[
    arr/.style={->, >=stealth, thick},
    sarr/.style={->, >=stealth, semithick},
    block/.style={draw, thick, rounded corners=3pt},
    ph/.style={font=\small\sffamily\bfseries},
    ann/.style={font=\small\sffamily, text=gray!65!black},
    ell/.style={font=\normalsize\sffamily, text=black},
    mc/.style={draw=gray!50, minimum size=0.55cm,
               font=\fontsize{6}{7}\selectfont\sffamily, inner sep=0},
    mhi/.style={mc, fill=orange!45},
    mmid/.style={mc, fill=orange!20},
    mlo/.style={mc, fill=blue!18},
    sc/.style={draw=gray!50, minimum size=0.55cm,
               font=\fontsize{6}{7}\selectfont\sffamily, inner sep=0},
    smean/.style={sc, fill=orange!30},
    sstd/.style={sc,  fill=blue!20},
    sturn/.style={sc,  fill=teal!22},
    sscal/.style={sc,  fill=purple!20},
]

\def\ryA{2.10}  
\def\ryB{1.30}  
\def\ryC{0.50}  
\def\ryD{-0.25} 
\def\hdrY{3.30} 
\def\colY{2.8} 


\node[ph] at (1.22, \hdrY) {Trajectories};

\node[ann] at (0.20, \colY) {$t{=}1$};
\node[ann] at (1.05, \colY) {$t{=}2$};
\node[ell] at (1.7, \colY) {$\cdots$};
\node[ann] at (2.3, \colY) {$t{=}T$};

\foreach \wi/\wy/\fc/\dc in {
        1/\ryA/green!25/green!55!black,
        2/\ryB/orange!22/orange!60!black,
        3/\ryC/violet!18/violet!55!black}{
    \node[ann] at (-0.38, \wy) {$\world_{\wi}$};
    \foreach \tx in {0.20, 1.05, 2.25}
        \node[block, fill=\fc, draw=\dc, minimum size=0.65cm] at (\tx, \wy) {};
    \node[ell] at (1.7, \wy) {$\cdots$};
}

\draw[dashed, black, line width=0.8pt, rounded corners=2pt]
    (0.65, 0.045) rectangle (1.45, 2.55);
\node[ann] at (1.05, -0.20) {one timestep};


\draw[arr] (2.60, \ryB) -- node[above=1pt, ann] {count alive} (4.3, \ryB);


\node[ph] at (5.5, \hdrY) {Species fractions};

\node[ann] at (5.20, \colY) {env};
\node[ann] at (5.75, \colY) {$c_1$};
\node[ann] at (6.30, \colY) {$c_2$};

\node[ann] at (4.60, \ryA) {$t{=}1$};
\node[ann] at (4.60, \ryB) {$t{=}2$};
\node[ell] at (4.60, 0.90) {$\vdots$};
\node[ann] at (4.60, \ryC) {$t{=}T$};

\node[mhi]  at (5.20, \ryA) {.55};
\node[mmid] at (5.75, \ryA) {.30};
\node[mlo]  at (6.30, \ryA) {.15};
\node[mmid] at (5.20, \ryB) {.40};
\node[mmid] at (5.75, \ryB) {.35};
\node[mmid] at (6.30, \ryB) {.25};
\node[mlo]  at (5.20, \ryC) {.20};
\node[mhi]  at (5.75, \ryC) {.45};
\node[mhi]  at (6.30, \ryC) {.35};


\draw[arr] (6.58, \ryB) -- (7.45, \ryB);
\node[ann, above=1pt] at (7.005, \ryB) {col};


\node[ph] at (8.7, \hdrY) {Features};

\node[ann, anchor=east] at (7.85, \ryA) {$\boldsymbol{\mu}$};
\node[ann, anchor=east] at (7.85, \ryB) {$\boldsymbol{\sigma}$};
\node[ann, anchor=east] at (7.85, \ryC) {$\boldsymbol{\delta}$};

\node[smean] at (8.15, \ryA) {.38};
\node[smean] at (8.70, \ryA) {.37};
\node[smean] at (9.25, \ryA) {.25};

\node[sstd] at (8.15, \ryB) {.14};
\node[sstd] at (8.70, \ryB) {.06};
\node[sstd] at (9.25, \ryB) {.10};

\node[sturn] at (8.15, \ryC) {.18};
\node[sturn] at (8.70, \ryC) {.08};
\node[sturn] at (9.25, \ryC) {.12};

\draw[dashed, gray!50, line width=0.5pt] (7.87, 0.15) -- (9.03, 0.15);

\node[ann, anchor=east] at (7.85, \ryD)
    {$H_{\mathrm{win}},\,\nu$};
\node[sscal] at (8.15, \ryD) {.87};
\node[sscal] at (8.70, \ryD) {.23};

\node[ann, align=left, anchor=west] at (9.00, \ryD)
    {\text{from argmax\,/\,$|\Delta\mathbf{a}|$}};


\draw[arr] (9.53, \ryB) -- (10.4, \ryB);
\node[ann, above=1pt] at (9.92, \ryB) {cat};
\node[ann, below=1pt] at (9.92, \ryB) {$\ell_2$};


\node[ph] at (11.30, \hdrY) {Descriptor};

\def\bx{10.45}
\def\by{\ryB}
\def\bh{0.55}

\fill[orange!30] (\bx,       \by-\bh/2) rectangle (\bx+0.5, \by+\bh/2);
\fill[blue!20]   (\bx+0.5,  \by-\bh/2) rectangle (\bx+1.0, \by+\bh/2);
\fill[teal!22]   (\bx+1.0,  \by-\bh/2) rectangle (\bx+1.5, \by+\bh/2);
\fill[purple!20] (\bx+1.5,  \by-\bh/2) rectangle (\bx+1.9, \by+\bh/2);
\draw[thick] (\bx, \by-\bh/2) rectangle (\bx+1.92, \by+\bh/2);

\node[ann] at (\bx+0.26,  \by+\bh/2+0.16) {$\boldsymbol{\mu}$};
\node[ann] at (\bx+0.78,  \by+\bh/2+0.16) {$\boldsymbol{\sigma}$};
\node[ann] at (\bx+1.30,  \by+\bh/2+0.16) {$\boldsymbol{\delta}$};
\node[ann] at (\bx+1.74,  \by+\bh/2+0.16) {$H,\nu$};

\node[ann] at (\bx+0.96, \by-\bh/2-0.22)
    {$\mathbf{d}_i \in \mathbb{R}^{3(N+1)+2}$};


\node[ph] at (15.0, \hdrY) {Archive novelty};

\node[ann] at (12.55, \ryA+0.20) {$k$-NN};

\foreach \dy/\alpha in {0.60/30, 0.0/50, -0.60/70} {
    \fill[gray!\alpha] (13.00, \ryB+\dy-0.20) rectangle (13.70, \ryB+\dy+0.20);
    \draw[gray!60]     (13.00, \ryB+\dy-0.20) rectangle (13.70, \ryB+\dy+0.20);
}
\node[ann, below=2pt] at (13.5, \ryC-0.10) {archive $\mathcal{A}$, FIFO buffer};

\foreach \dy in {0.60, 0.0, -0.60} {
    \draw[sarr, gray!55, dashed]
        (\bx+1.92, \by) -- (13.00, \ryB+\dy);
}

\draw[arr] (\bx+1.93, \by) -- (13, \by);


\node[block, fill=gray!1, draw=gray!40,
      minimum width=3.6cm, minimum height=1.00cm] at (15.80, \ryB) {};
\node[font=\small\sffamily] at (15.80, \ryB)
    {$\mathcal{N}_i = \dfrac{1}{k}\displaystyle\sum_{j \in \mathrm{kNN}} \|\mathbf{d}_i - \mathbf{d}_j\|_2$};

\end{tikzpicture}

}
\caption{Handcrafted behavior descriptor and novelty score.
Each world is rolled out; the first $N+1$ channels of the
trajectory (alive mass per entity) are spatially summed and row-normalised to produce species-fraction trajectories. The per-entity statistics are extracted: mean occupancy $\boldsymbol{\mu}$, temporal standard deviation $\boldsymbol{\sigma}$, and turnover $\boldsymbol{\delta}$. We use a $k=8$ for the kNN.}

\label{fig:handcrafted_descriptor}
\end{figure*}

%
%
%

\begin{figure}[ht]
    \centering
    \resizebox{\linewidth}{!}{%
    \begin{tikzpicture}[
        arr/.style={->, >=stealth, thick},
        block/.style={draw, thick, rounded corners=3pt},
        ph/.style={font=\small\sffamily\bfseries},
        ann/.style={font=\small\sffamily, text=gray!65!black},
        ell/.style={font=\normalsize\sffamily, text=black},
        cell/.style={draw, minimum size=0.65cm, font=\scriptsize\sffamily},
        hicell/.style={cell, fill=red!22},
        locell/.style={cell, fill=blue!18},
        diagcell/.style={cell, fill=gray!18},
    ]


    \node[ph] at (1.22, 3.0) {Trajectories};   

    \node[ann] at (0.20, 2.5) {$t{=}1$};
    \node[ann] at (1.05, 2.5) {$t{=}2$};
    \node[ell] at (1.73, 2.5) {$\cdots$};      
    \node[ann] at (2.3, 2.5) {$t{=}T$};

    \foreach \wi/\wy/\fc/\dc in {
            1/1.80/green!25/green!55!black,
            2/1.10/orange!22/orange!60!black,
            3/0.40/violet!18/violet!55!black}{
        \node[ann] at (-0.38, \wy) {$\world_{\wi}$};
        \foreach \tx in {0.20, 1.05, 2.3}
            \node[block, fill=\fc, draw=\dc, minimum size=0.65cm] at (\tx, \wy) {};
        \node[ell] at (1.73, \wy) {$\cdots$};   
    }

    \draw[dashed, black, line width=0.8pt, rounded corners=2pt]
        (0.65, -0.05) rectangle (1.45, 2.25);
    \node[ann, font=\large] at (1.05, -0.3) {one timestep};


    \node[ph] at (3.75, 3.0) {Embed};

    \draw[arr] (2.65, 1.10) -- (3.15, 1.10);

    \node[block, fill=teal!12, draw=teal!60!black,
          minimum width=1.0cm, minimum height=1.90cm] at (3.65, 1.10) {};
    \node[ph, text=teal!70!black, rotate=90] at (3.65, 1.10) {DINOv2};
    \node[ann, font=\large] at (3.70, -0.30) {$\mathbf{z}_i^t \in \mathbb{R}^{768}$};

    %

    \node[ph] at (5.7, 3.0) {Distance};

    \draw[arr] (4.18, 1.10) -- (6.89, 1.10);
    \node[ann, above=2pt, font=\large] at (5.58, 1.10)
        {$d_{ij}^t = 1 - \langle \mathbf{z}_{i}^{t}, \mathbf{z}_{j}^{t} \rangle$};
    \foreach \wi/\x in {1/7.80, 2/8.40, 3/9.0}
        \node[ann] at (\x, 2.35) {$\world_{\wi}$};

    \foreach \wi/\y in {1/1.82, 2/1.18, 3/0.54}
        \node[ann] at (7.2, \y) {$\world_{\wi}$};

    \node[diagcell] at (7.80,1.82) {$-$};
    \node[hicell]   at (8.40,1.82) {$.82$};   
    \node[hicell]   at (9.0,1.82) {$.77$};   

    \node[hicell]   at (7.80,1.18) {$.82$};   
    \node[diagcell] at (8.40,1.18) {$-$};
    \node[locell]   at (9.0,1.18) {$.29$};   

    \node[hicell]   at (7.80,0.54) {$.77$};   
    \node[locell]   at (8.40,0.54) {$.29$};   
    \node[diagcell] at (9.0,0.54) {$-$};


    \node[ph] at (9.5, 3.0) {Row median};

    \foreach \y/\val in {1.82/.80, 1.18/.56, 0.54/.53}{
        \draw[arr] (9.35, \y) -- (9.85, \y);
        \node[ann, right, font=\small] at (9.75, \y+0.03) {$\operatorname{med}=\val$};
    }

    \draw[arr, gray!55] (10.1, 0.30) -- (10.1, -0.4);
    \node[ann, right=2pt, font=\large] at (10.1, -0.05) {avg $\forall\, t$};

    \node[block, fill=gray!1, draw=gray!40,
          minimum width=4.6cm, minimum height=1.3cm] at (9.50, -1.1) {};
    \node[font=\large] at (9.50, -1.1)
        {$\mathcal{D}_i = \dfrac{1}{T}\displaystyle\sum_{t=1}^{T}
          \operatorname{med}_{j \neq i}\, d_{ij}^t$};

    \end{tikzpicture}
    }
    \caption{Visual diversity score. At each timestep $t$, frames from all $P$ worlds are embedded using a DINOv2 encoder. We compute the cosine distance matrix with masked diagonals to score each world through its median distance to all other worlds, averaged across the rollout horizon.}
    \label{fig:dino_scoring}
    \vspace{-5pt}
\end{figure}
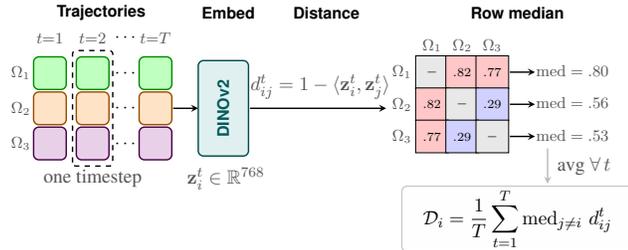

\subsection{Population State and World Rollouts}

We maintain a population of $P$ worlds. A world encompasses the full state of an ongoing adaptive ecosystem. Concretely, each world stores:  
(i) the parameters of all resident NCAs;  
(ii) their optimizer states;  
(iii) the mutable world state, including the grid contents, rollout counters, and any auxiliary state needed by the simulator; and  
(iv) learning hyperparameters controlling within-world adaptation, such as learning rate, batch size, and update frequency.

At each meta-iteration, every world is rolled out for a fixed horizon of $T_{\mathrm{world}}$ inner steps. During this rollout, cellular updates and gradient-based learning remain interleaved, so the world's agents continue adapting while interacting on the grid. This produces a trajectory $\mathcal{T} = (X^1,\dots,X^{T_{\mathrm{world}}})$, which we utilize as the object for evaluating each world's novelty score. Scoring full trajectories rather than terminal frames is important because worlds with similar endpoints may differ greatly in coexistence ratios between different NCAs across time, transient competition or point of collapse.

\subsection{Composite Scoring Function}

We score each world $\world_i$, $i\in\{ 1,\dots, P \}$, with a composite scalar function $\mathcal{F}_i = \mathcal{N}_i + \mathcal{D}_i$, where $\mathcal{N}_i$ is an archive-based novelty score computed from a handcrafted behavioral descriptor, and $\mathcal{D}_i$ is a population-level visual diversity score computed from image embeddings. These terms are complementary: $\mathcal{N}_i$ captures how distinctly a world $\world_i$ behaves relative to previously discovered ones, while $\mathcal{D}_i$ captures how much the world's individual timesteps differ from those of other worlds currently being trained in the population.

\subsubsection{Behavioral Descriptor and Archive Novelty.}
The archive novelty term is based on handcrafted features of substrate dynamics as illustrated in~\autoref{fig:handcrafted_descriptor}. 
Instead of comparing high-dimensional grid states directly, we summarize each trajectory using statistics of fundamental concepts in ecology like species occupancy and change~\citep{magurran2004measuring}.
For each inner time step, we compute the fraction of total alive mass belonging to each entity, i.e., each of the $N$ agents and the environment channel. We extract the following feature groups from the timeseries of each species' fractions:
\begin{enumerate}[leftmargin=*]\setlength{\itemsep}{0pt}\setlength{\parskip}{2pt}\setlength{\parsep}{0pt}
    \item \textbf{Mean occupancy} $\mu$: the average fraction of alive mass assigned to each entity over the rollout.
    \item \textbf{Temporal standard deviation} $\sigma$: the variability of each entity's occupancy over time.
    \item \textbf{Mean frame-to-frame occupancy change} $\delta$: the mean rate at which a grid cell changes from one agent (NCA) to another.
    \item \textbf{Winner-map entropy} $H$: at each spatial location, we identify the dominant alive channel and compute the entropy of the winner's distributions across space and time to summarize how mixed or monopolized the world is.
    \item \textbf{Mean absolute alive-mass change} $\nu$: the average frame-to-frame alive mask change.
\end{enumerate}
These statistics are concatenated and $\ell_2$-normalized to form the feature vector
$d \in \mathbb{R}^{3(N+1)+2}$.
This descriptor is more stable under small perturbations in the world state compared to the high-dimensional original frames and ensures that separate ecological regimes such as collapse, oscillation, dominance, stable or volatile coexistence have distinct descriptors.

We store past descriptors in a bounded FIFO archive $\mathcal{A}$~\citep{pugh2016qdmaps,lehman2011novelty}. After each meta-iteration, the descriptors of the top-$m$ ($m=2$) worlds under the current novelty score are appended to the archive; if the archive exceeds its capacity $A_{\max}$, the oldest entries are removed first. Optionally, the archive may be cleared every $R$ meta-iterations to reset the novelty signal once too many previously discovered regimes have accumulated.
Given the archive, the novelty score $\mathcal{N}_i$ of a world $\world_i$ is the mean Euclidean distance from its descriptor to the $k$ ($k=8$) nearest neighbors in $\mathcal{A}$~\citep{lehman2011novelty}, making it interesting when it differs from what has already been discovered.

\begin{figure*}[ht]
\centering

\begin{minipage}{0.325\linewidth}
    \centering
    {\scriptsize $\mathtt{meta\text{-}iter\ 20}$}
    \vspace{0.3em}
    
    \includegraphics[width=0.32\linewidth]{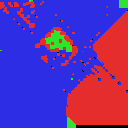}
    \includegraphics[width=0.32\linewidth]{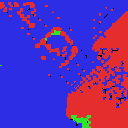}
    \includegraphics[width=0.32\linewidth]{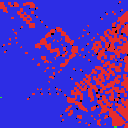}
    
\end{minipage}
\hfill
\begin{minipage}{0.325\linewidth}
    \centering
    {\scriptsize $\mathtt{meta\text{-}iter\ 125}$}
    \vspace{0.3em}
    
    \includegraphics[width=0.32\linewidth]{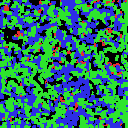}
    \includegraphics[width=0.32\linewidth]{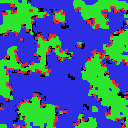}
    \includegraphics[width=0.32\linewidth]{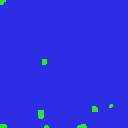}
    
\end{minipage}
\hfill
\begin{minipage}{0.325\linewidth}
    \centering
    {\scriptsize $\mathtt{meta\text{-}iter\ 370}$}
    \vspace{0.3em}

    \includegraphics[width=0.32\linewidth]{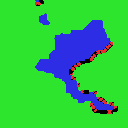}
    \includegraphics[width=0.32\linewidth]{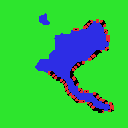}
    \includegraphics[width=0.32\linewidth]{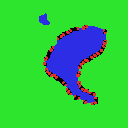}
    
\end{minipage}

    \caption{Shooters, Archipelago and Ant locomotion emerging from PBT-NCA (3 NCAs) at meta-iteration 20, 125 and 370.}
    \label{fig:pd-nca-ants}
\end{figure*}

\begin{figure*}[ht]
\centering
\resizebox{\linewidth}{!}{
    \includegraphics[width=0.1\linewidth]{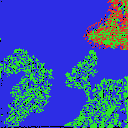}
    \includegraphics[width=0.1\linewidth]{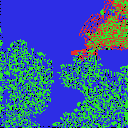}
    \includegraphics[width=0.1\linewidth]{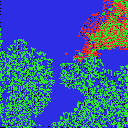}
    \includegraphics[width=0.1\linewidth]{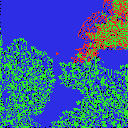}
    \includegraphics[width=0.1\linewidth]{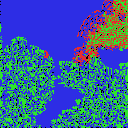}
    \includegraphics[width=0.1\linewidth]{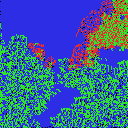}
    \includegraphics[width=0.1\linewidth]{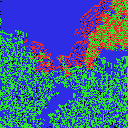}
    \includegraphics[width=0.1\linewidth]{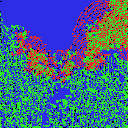}
    \includegraphics[width=0.1\linewidth]{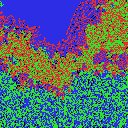}
}
\caption{Self-replication emerging in PBT-NCA (3 NCA agents). A macroscopic entity (red agent) emits a small cluster of cells across the substrate, executing spatial colonization via a decentralized replicating strategy.}
\label{fig:colonize-collab}
\end{figure*}

\vspace{-7pt}
\subsubsection{Visual Diversity with DINOv2.}

The archive descriptor captures relevant ecological statistics but ignores many visual properties of the rollout, such as morphology, spatial arrangement, and color pattern. We draw inspiration from ASAL \citep{kumar2025automating} to define a second term (\autoref{fig:dino_scoring}) based on a frozen DINOv2 vision encoder~\citep{oquab2023dinov2} to complement it.
For each world, we extract all the frames from its trajectory and encode them into a DINOv2 embedding. Let $\mathbf{z}_i^t$ denote the embedding of world $\world_i$ at sampled frame $t$. For each frame, we compare world $i$ only to the other worlds $j \neq i$ at the same sampled time point and compute cosine distances: $d_{ij}^t = 1 - \langle \mathbf{z}_{i}^{t}, \mathbf{z}_{j}^{t} \rangle$.
To ensure robustness to outliers, the per-frame diversity of world $\world_i$ is defined as the median pairwise distance across the population. Averaging over sampled frames gives the final diversity score $\mathcal{D}_i = 1/T \sum_{t=1}^{T} \operatorname{median}_{j \neq i}( d_{ij}^{t})$, where $T$ is the number of sampled frames. Intuitively, $\mathcal{D}_i$ is high when a world produces visual structures that the rest of the population is not currently producing. Since DINOv2 provides a rich perceptual representation without any task-specific labels, this term rewards novel spatial texture or morphology even when the handcrafted descriptor is similar across the population of worlds \citep{Zhang_2018_CVPR,niklasson2021self-organising}.

\vspace{-5pt}

\begin{table}[h]
\centering
\caption{PBT hyperparameter search space. $^\dagger$ means that the hyperparameter is in the extended search space.}
\label{tab:hyperparams}
\resizebox{\linewidth}{!}{
\begin{tabular}{l l l c c}
\toprule
HP & Type & Range & Log & Default \\
\midrule
\texttt{learning\_rate} & float & $[10^{-6},\,1.0]$ & Yes & $3\times10^{-4}$ \\
\texttt{batch\_size} & int & $[1,\,8]$ & No & 8 \\
\texttt{steps\_per\_update} & int & $[1,\,64]$ & No & 4 \\
\midrule
$^\dagger$\texttt{softmax\_temp} & float & $[0.05,\,5.0]$ & Yes & 1.0 \\
$^\dagger$\texttt{per\_hid\_upd}  & float & $[0.05,\,1.0]$ & No & 1.0 \\
\bottomrule
\end{tabular}
}
\end{table}

\subsection{Exploit--Explore}

Once all worlds have been rolled out and scored, we rank the population by the composite score function $\mathcal{F}$. The top-$m$ behavioral descriptors are inserted into the archive as described above. Every $K$ meta-iterations, we then perform a population-based exploit--explore~\citep{jaderberg2017pbt,liu2019emergent}. We set the number of worlds to replace to $n_{\mathrm{replace}} = \mathrm{round}(\rho P)$,
where $\rho \in (0,1)$ is the replacement fraction and $P$ is the population size.
The $n_{\mathrm{replace}}$ lowest-scoring worlds are discarded and replaced by higher-scoring worlds. For each replacement, a parent is sampled uniformly from the elite subset, and the child is generated in four stages:
\begin{enumerate}[leftmargin=*]\setlength{\itemsep}{0pt}\setlength{\parskip}{2pt}\setlength{\parsep}{0pt}
    \item \textbf{Copy.} The parent world is deep-copied into the child. This includes neural-network parameters, optimizer state, world state, and hyperparameters. The inheritance is therefore Lamarckian: descendants do not restart from random initialization but continue from a partially developed ecological and optimization state.
    \item \textbf{Crossover.} Each hyperparameter is independently crossed over. With probability $P_{cross}$ ($0.5$ in our experiments), the copied parent value is retained; otherwise, the child reverts to its own pre-copy value. Thus, the search mixes successful inherited settings with previously held settings.
    \item \textbf{Mutation.} Each hyperparameter is perturbed independently with probability $P_{pert}$ ($0.1$ in our experiments). The perturbation multiplies the current value by either $1.2$ or $0.8$, chosen with equal probability, and clips the result to its allowed range. Integer hyperparameters are rounded after scaling.
    \item \textbf{Weight perturbation.} Finally, independent Gaussian noise is added to all trainable neural network parameters. This injects local parametric variation without destroying the inherited world context.
\end{enumerate}
Steps 1 to 3 are directly adapted from \citet{jaderberg2017pbt} and \citet{liu2019emergent}.
The exploit phase (step 1) amplifies lineages that currently produce novel and diverse trajectories, whereas the explore phase (steps 2-4) evaluates nearby dynamical possibilities by perturbing both hyperparameters and weights of the worlds in the population. 
Overall, the method couples two time scales. Within a rollout, gradient-based adaptation shapes the behavior of each world. Across meta-iterations, population-based selection decides which adaptive ecological regimes survive, diversify, and continue developing. This combination is what allows the system to search complex dynamics in an ever-diversifying population of PD-NCAs, which would be difficult to discover with a non-dynamical outer meta-optimization loop.

\section{Experiments and Simulations}
Our experiments aim to quantify the extent to which PBT-NCA sustains the continuous discovery of emerging complex dynamics and behaviours in the population. In all following experiments we use the hyperparameter space shown in Table~\ref{tab:hyperparams}. We set the population size to $P=30$, meta-iterations to $T = 500$, world horizon to $T_{world}=12$, explore-exploit interval to $K=5$, and the replacement factor to $\rho = 0.25$. We use Adam~\citep{kingma2015adam} to optimize the NCA weights with the sampled/mutated learning rate.
Unless stated otherwise, we use 3 NCA agents in each world. Figure~\ref{fig:pd-nca-ants} and \ref{fig:colonize-collab} demonstrate several complex dynamics, such as directed projectile emission and colonization from stable territorial clusters, as well as agents following learned territorial trails---all emerging purely from local NCA competition.

\begin{figure}[t]
\centering
\begin{minipage}{0.325\linewidth}
    \centering
    {\scriptsize $\mathtt{iter\ 100}$}
    \vspace{0.3em}
    
    \includegraphics[width=0.9\linewidth]{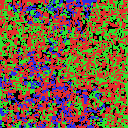}
\end{minipage}
\hfill
\begin{minipage}{0.325\linewidth}
    \centering
    {\scriptsize $\mathtt{iter\ 50000}$}
    \vspace{0.3em}
    
    \includegraphics[width=0.9\linewidth]{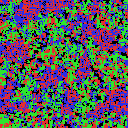}
\end{minipage}
\hfill
\begin{minipage}{0.325\linewidth}
    \centering
    {\scriptsize $\mathtt{iter\ 100000}$}
    \vspace{0.3em}
    
    \includegraphics[width=0.9\linewidth]{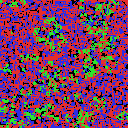}
\end{minipage}   
    \caption{Original PD-NCA (3 NCA agents) frames on iterations 100, 50000 and 100000.}
    \label{fig:pd-nca-baseline}
\end{figure}

\begin{figure}[t]
\centering
\begin{minipage}{0.325\linewidth}
    \centering
    {\scriptsize $\mathtt{meta\text{-}iter\ 145}$}
    \vspace{0.3em}
    
    \includegraphics[width=0.9\linewidth]{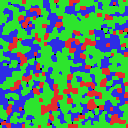}
\end{minipage}
\hfill
\begin{minipage}{0.325\linewidth}
    \centering
    {\scriptsize $\mathtt{meta\text{-}iter\ 485}$}
    \vspace{0.3em}
    
    \includegraphics[width=0.9\linewidth]{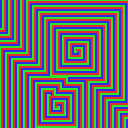}
\end{minipage}
\hfill
\begin{minipage}{0.325\linewidth}
    \centering
    {\scriptsize $\mathtt{meta\text{-}iter\ 495}$}
    \vspace{0.3em}
    
    \includegraphics[width=0.9\linewidth]{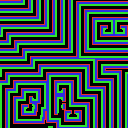}
\end{minipage}   
    \caption{Random Search (3 NCA agents) frames on meta-iterations 145, 485 and 495.}
    \label{fig:rs-baseline}
    \vspace{-1em}
\end{figure}

\subsubsection{Comparison to Baselines.}
We compare our PBT-NCA against two other configurations:
\begin{enumerate}\setlength{\itemsep}{0pt}\setlength{\parskip}{2pt}\setlength{\parsep}{0pt}
    \item \textbf{Fixed-Hyperparameters PD-NCA.} The original implementation from \citet{pdnca2025} using the default fixed hyperparameters from the original implementation.
    \item \textbf{Random Search (RS).} Hyperparameters are sampled uniformly from the PBT-NCA search space (Table~\ref{tab:hyperparams}).
\end{enumerate}
\vspace{-0.1cm}
To match the compute budget of PBT-NCA, we use the same population size, $P=30$, for RS. Each world was trained (rolled out) for a total of 6000 iterations ($T = 500$ meta-iterations $\times$ $T_{world} = 12$ iterations). RS discovers some interesting regular cyclic dynamics (Figure~\ref{fig:rs-baseline}), whereas PD-NCA exhibits high entropy noise (Figure~\ref{fig:pd-nca-baseline}).

\begin{figure}[t]
    \centering
    \includegraphics[width=0.99\linewidth]{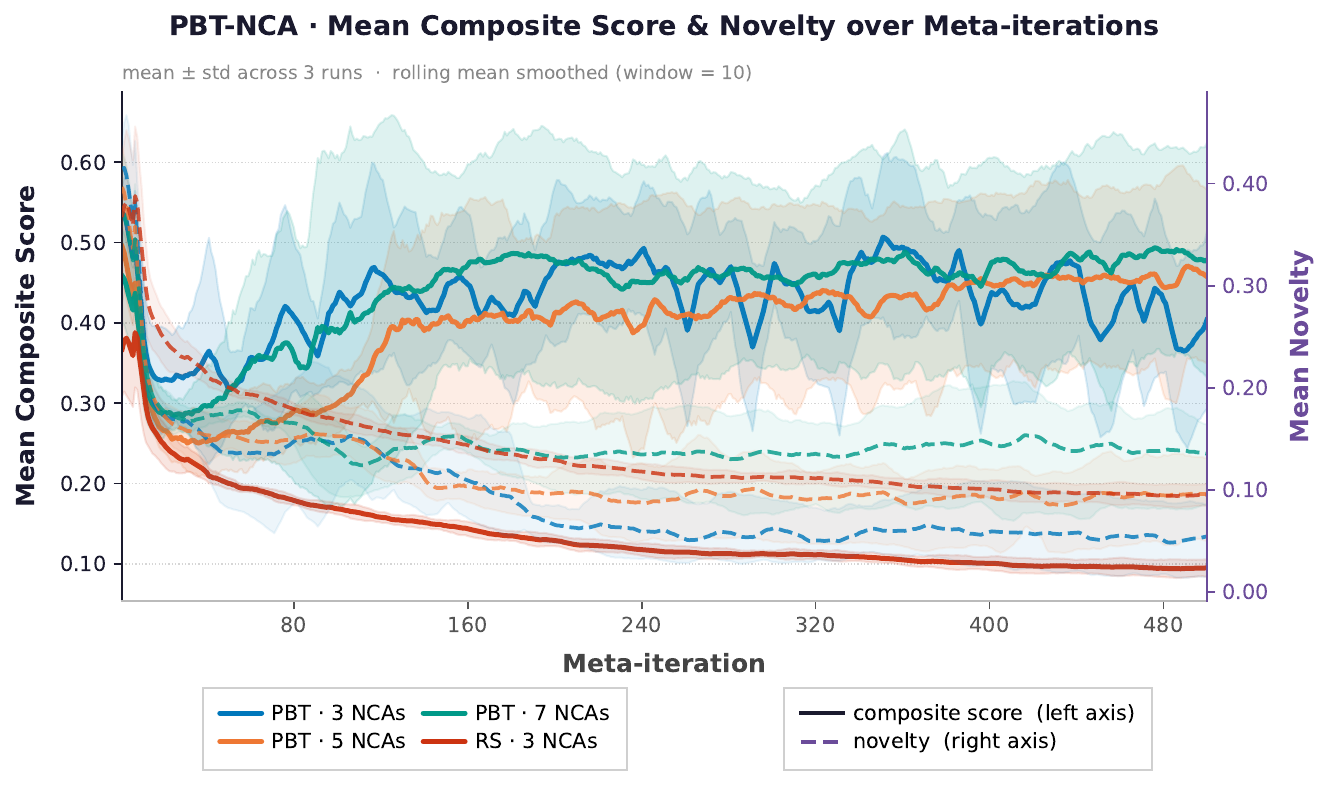}\\
    \includegraphics[width=0.99\linewidth]{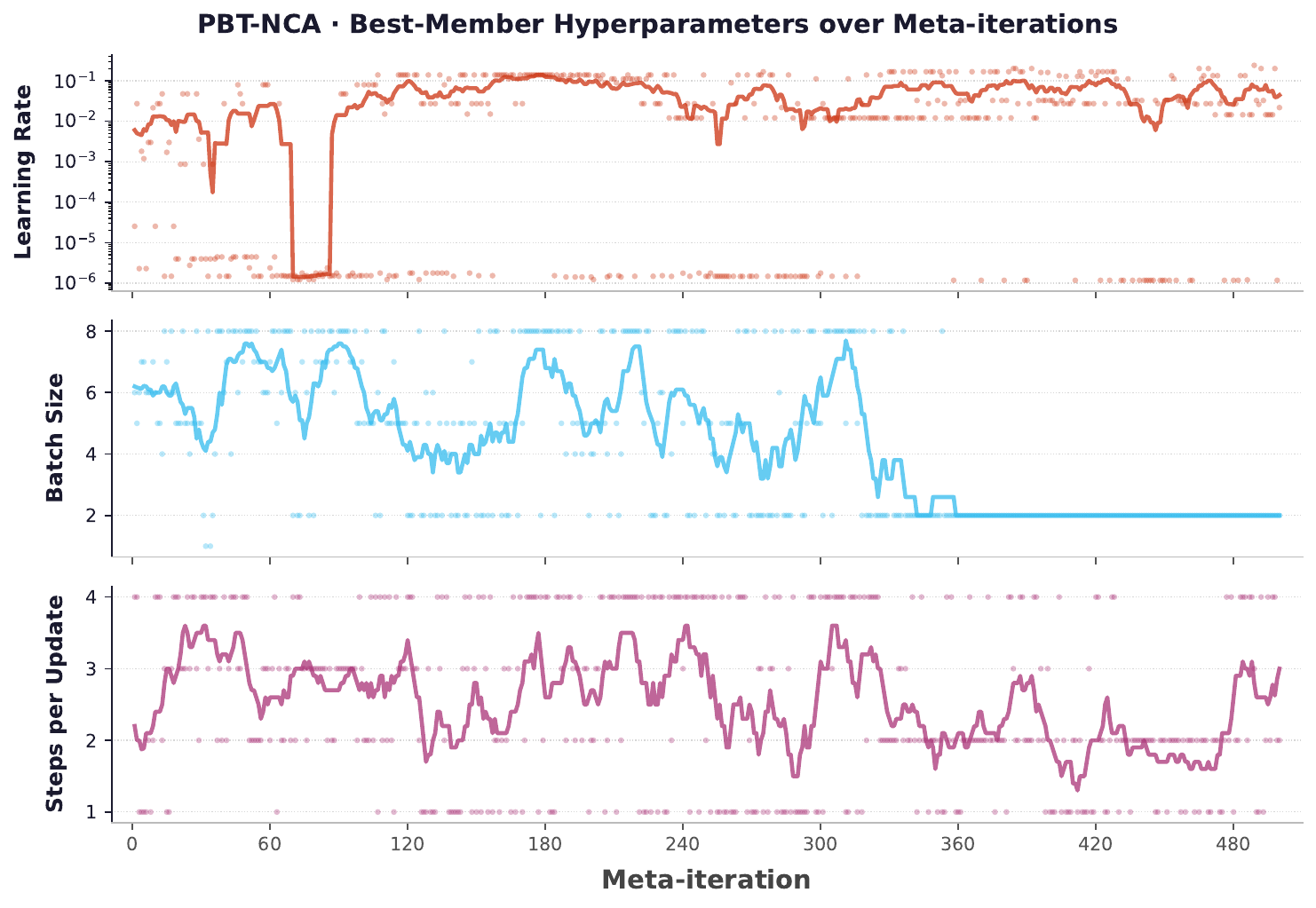}

    \vspace{-1em}
    \caption{\textbf{Top figure:} Mean novelty and composite score function over the number of meta-iterations. \textbf{Bottom figure:} Sampled hyperparameters over the number of meta-iterations for the best PBT-NCA (7 NCA agents).}
    \label{fig:fitness_hyperparams}
    \vspace{-1em}
\end{figure}

\begin{figure*}[t]
    \centering
    
    \resizebox{0.99\linewidth}{!}{
    \raisebox{0.00\linewidth}{\rotatebox{90}{\scriptsize $\mathtt{meta-iter\ 145}$}}\hspace{0.5em}
    \includegraphics[width=0.11\linewidth]{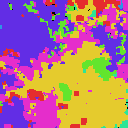}
    \includegraphics[width=0.11\linewidth]{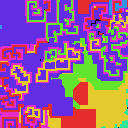}
    \includegraphics[width=0.11\linewidth]{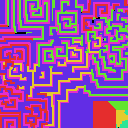}
    \includegraphics[width=0.11\linewidth]{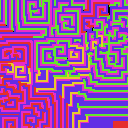}
    \includegraphics[width=0.11\linewidth]{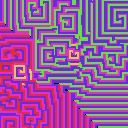}
    \includegraphics[width=0.11\linewidth]{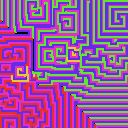}
    \includegraphics[width=0.11\linewidth]{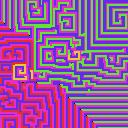}
    \includegraphics[width=0.11\linewidth]{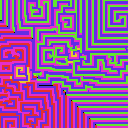}
    }
    
    \resizebox{0.99\linewidth}{!}{
    \raisebox{0.00\linewidth}{\rotatebox{90}{\scriptsize $\mathtt{meta-iter\ 230}$}}\hspace{0.5em}
    \includegraphics[width=0.11\linewidth]{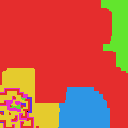}
    \includegraphics[width=0.11\linewidth]{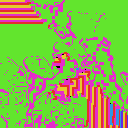}
    \includegraphics[width=0.11\linewidth]{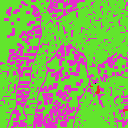}
    \includegraphics[width=0.11\linewidth]{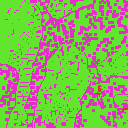}
    \includegraphics[width=0.11\linewidth]{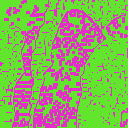}
    \includegraphics[width=0.11\linewidth]{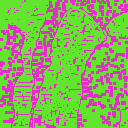}
    \includegraphics[width=0.11\linewidth]{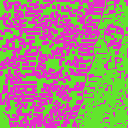}
    \includegraphics[width=0.11\linewidth]{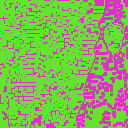}
    }

    \resizebox{0.99\linewidth}{!}{
    \raisebox{0.00\linewidth}{\rotatebox{90}{\scriptsize $\mathtt{meta-iter\ 310}$}}\hspace{0.5em}
    \includegraphics[width=0.11\linewidth]{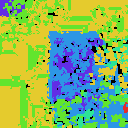}
    \includegraphics[width=0.11\linewidth]{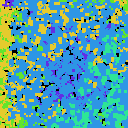}
    \includegraphics[width=0.11\linewidth]{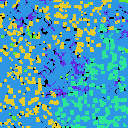}
    \includegraphics[width=0.11\linewidth]{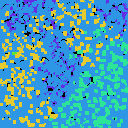}
    \includegraphics[width=0.11\linewidth]{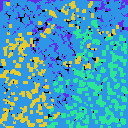}
    \includegraphics[width=0.11\linewidth]{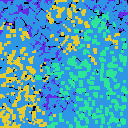}
    \includegraphics[width=0.11\linewidth]{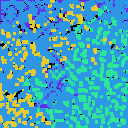}
    \includegraphics[width=0.11\linewidth]{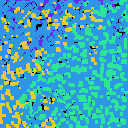}
    }

    \resizebox{0.99\linewidth}{!}{
    \raisebox{0.00\linewidth}{\rotatebox{90}{\scriptsize $\mathtt{meta-iter\ 395}$}}\hspace{0.5em}
    \includegraphics[width=0.11\linewidth]{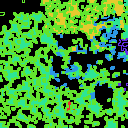}
    \includegraphics[width=0.11\linewidth]{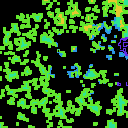}
    \includegraphics[width=0.11\linewidth]{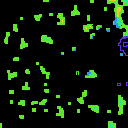}
    \includegraphics[width=0.11\linewidth]{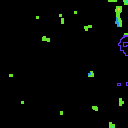}
    \includegraphics[width=0.11\linewidth]{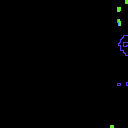}
    \includegraphics[width=0.11\linewidth]{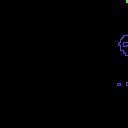}
    \includegraphics[width=0.11\linewidth]{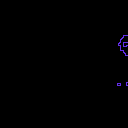}
    \includegraphics[width=0.11\linewidth]{figures/frames/pbt-nca-n7-2/alien_006.png}
    }

    \resizebox{0.990\linewidth}{!}{
    \raisebox{0.00\linewidth}{\rotatebox{90}{\scriptsize $\mathtt{meta-iter\ 460}$}}\hspace{0.5em}
    \includegraphics[width=0.11\linewidth]{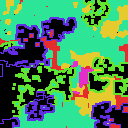}
    \includegraphics[width=0.11\linewidth]{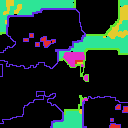}
    \includegraphics[width=0.11\linewidth]{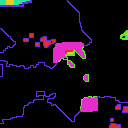}
    \includegraphics[width=0.11\linewidth]{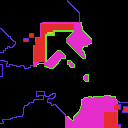}
    \includegraphics[width=0.11\linewidth]{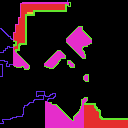}
    \includegraphics[width=0.11\linewidth]{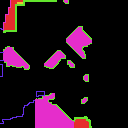}
    \includegraphics[width=0.11\linewidth]{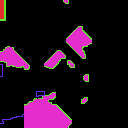}
    \includegraphics[width=0.11\linewidth]{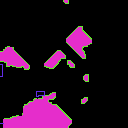}
    }

    \caption{Emergent dynamics at meta-iteration 145, 230, 310, 395 and 495 of PBT-NCA (7 NCA agents), demonstrating the continuous novelty generation, a sign of open-endedness. We discover novel entities, such as spirals (145), amoeba (230), pirate world (archipelago$+$shooters) (310), alien (395) and glider (460).}
    \label{fig:pbt-nca-emergent}
\end{figure*}

\begin{figure*}[t]
\centering

\begin{minipage}{0.325\linewidth}
    \centering
    {\scriptsize $\mathtt{meta\text{-}iter\ 20}$}
    \vspace{0.3em}

    \includegraphics[width=0.32\linewidth]{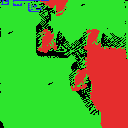}\hfil
    \includegraphics[width=0.32\linewidth]{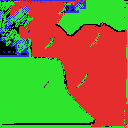}\hfil
    \includegraphics[width=0.32\linewidth]{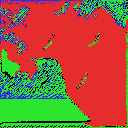}
    
\end{minipage}
\hfill
\begin{minipage}{0.325\linewidth}
    \centering
    {\scriptsize $\mathtt{meta\text{-}iter\ 95}$}
    \vspace{0.3em}
    
    \includegraphics[width=0.32\linewidth]{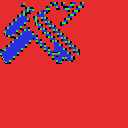}\hfil
    \includegraphics[width=0.32\linewidth]{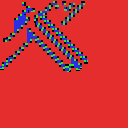}\hfil
    \includegraphics[width=0.32\linewidth]{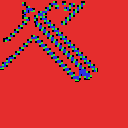}
    
\end{minipage}
\hfill
\begin{minipage}{0.325\linewidth}
    \centering
    {\scriptsize $\mathtt{meta\text{-}iter\ 125}$}
    \vspace{0.3em}

    \includegraphics[width=0.32\linewidth]{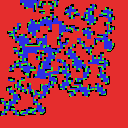}\hfil
    \includegraphics[width=0.32\linewidth]{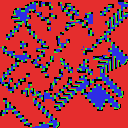}\hfil
    \includegraphics[width=0.32\linewidth]{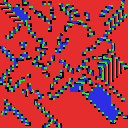}
    
\end{minipage}

\caption{Rigid, circuit-like geometric patterns discovered with PBT-NCA (3 NCA agents) on the extended search space. We notice the emergence of spaceships, gliders and information propagation, as discovered in continuous or discrete CA.}
\label{fig:pd-nca-gliders-shooters}
\vspace{-12pt}
\end{figure*}

\subsection{Novelty, Open-Endedness, and Scalability}
To evaluate population-level performance, we track the mean composite score, $\mathcal{\bar{F}} =1/P \sum_{i=1}^P \mathcal{F}_i$, across all members of the population. Figure~\ref{fig:fitness_hyperparams} (top) illustrates the progression of $\mathcal{\bar{F}}$ over the PBT-NCA meta-iterations (mean and standard deviation from three independent PBT-NCA runs) for worlds containing 3, 5, and 7 NCA agents. Across all configurations, an initial decline is followed by a steady increase in the composite score, suggesting the emergence of a diverse population of PBT-NCA substrates capable of continually generating novel dynamics (see Figures~\ref{fig:frames_fitness_7} and \ref{fig:pbt-nca-emergent}). Furthermore, to isolate the individual components of the composite score, we plot the \textit{mean novelty}, $\mathcal{\bar{N}}$, in the same figure. 
We observe that worlds scaled to support a larger number of NCAs successfully retain higher levels of novelty in the later stages of training. We observe a strong selection pressure toward higher learning rates and smaller batch sizes (Figure~\ref{fig:fitness_hyperparams}, bottom). By injecting greater gradient noise and accelerating parameter updates, this combination prevents worlds from collapsing into stagnant equilibria, directly supporting our novelty-seeking objective.
Furthermore, when granted access to the extended search space (\autoref{fig:pd-nca-gliders-shooters}), PBT-NCA discovers qualitatively distinct dynamics, abandoning fluid, organic morphologies in favor of rigid, circuit-like geometric expansions. This demonstrates that PBT-NCA can effectively scale also along the hyperparameter dimensions to unlock entirely new classes of emergent behavior.

\subsection{Edge-of-Chaos Analysis}
\label{sec:edge-of-chaos}

To quantify whether the emergent dynamics occupy the \emph{edge of
chaos}---the narrow regime between frozen order and structureless
randomness where complex behaviour arises~\citep{langton1990,kauffman1993origins}---we
track two complementary metrics, similar to the ones used in \citet{dolson19}, over PBT-NCA with 7 agents
across 500~meta-steps.

\begin{figure}[ht]
    \centering
    \includegraphics[width=0.99\linewidth]{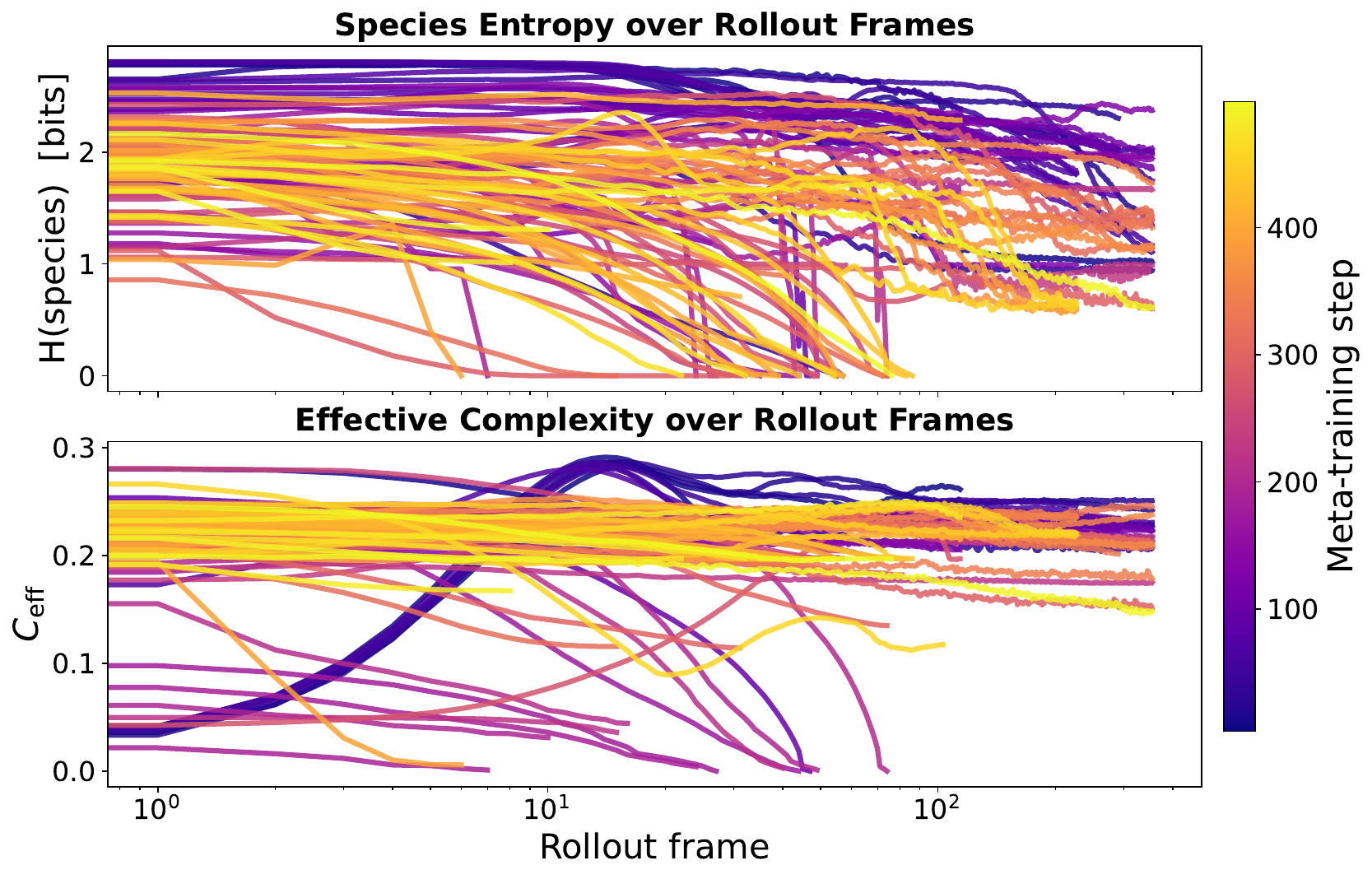} \\
    \includegraphics[width=0.99\linewidth]{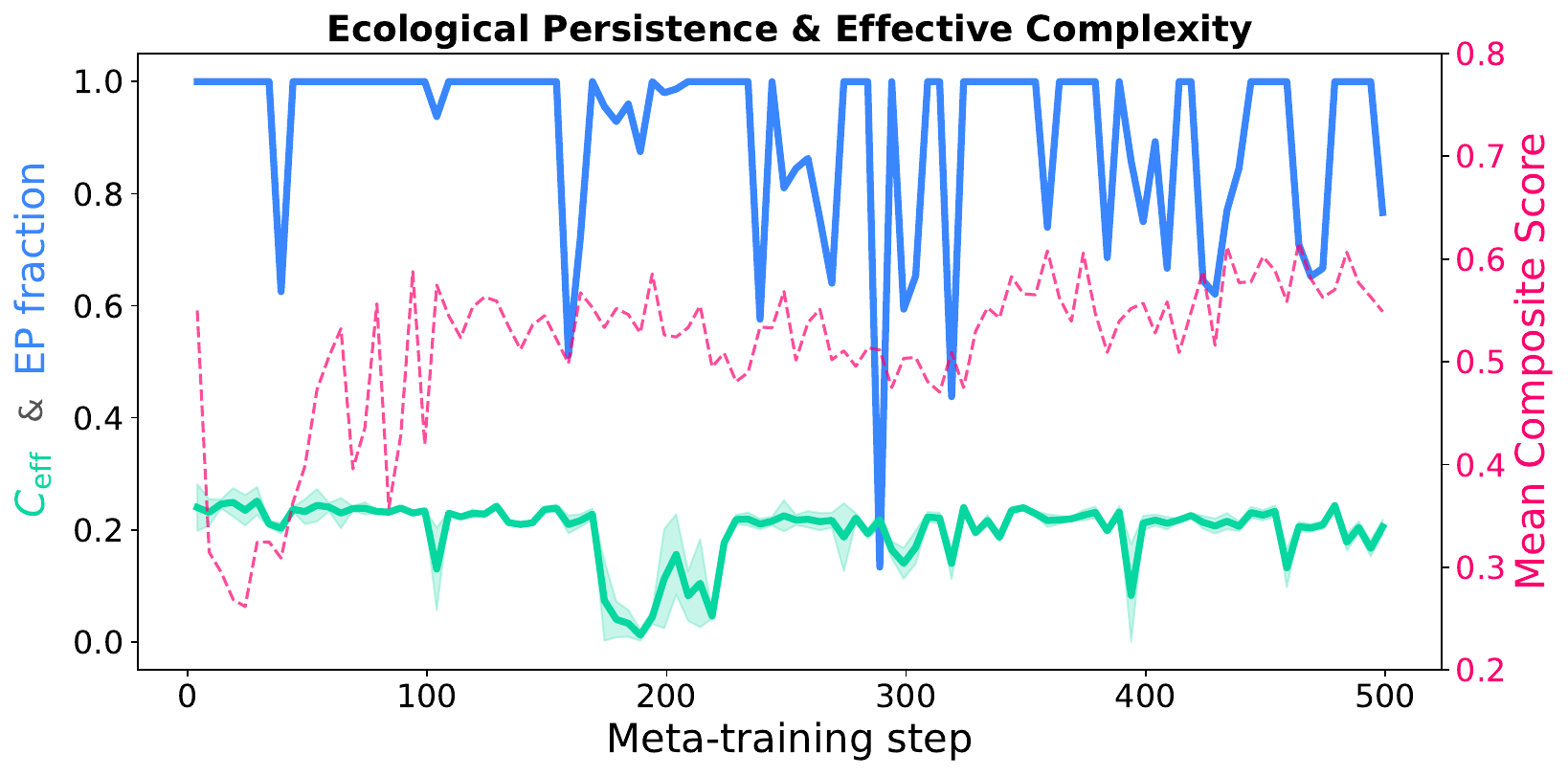}
    \caption{Ecological Persistence (EP) and Effective Complexity throughout meta-iterations and individual rollouts of PBT-NCA (7 NCA agents) worlds.}
    \label{fig:chaos_analysis}
    \vspace{-12pt}
\end{figure}

\paragraph{Ecological Persistence.}
Let $H^t$ be the per-pixel agent's aliveness distribution entropy at time~$t$. We define $EP = |\{t : H^t>\varepsilon\}|/T$ with $\varepsilon=0.1\,\log_2 N$, measuring the fraction of frames that sustain multi-agent coexistence. High $EP$ means the system avoids both extinction and monocultures.

\paragraph{Effective Complexity.}
Following \citet{gellmann1996}, we define
$\mathcal{C}_{\mathrm{eff}}^{t} = H(\mathrm{grid}_t)\times
(1-L_{\mathrm{Z77}}(\mathrm{grid}_t))$, where $H(\mathrm{grid}_t)$ is
the normalised spatial entropy and $L_{\mathrm{Z77}}$ is
compressibility ratio.  This product vanishes for both pure order
($H\!\approx\!0$) and pure noise ($L_{\mathrm{Z77}}\!\approx\!1$), and
peaks only when patterns are simultaneously varied \emph{and}
structurally compressible, the key characteristic of the \textit{edge of chaos}.

\paragraph{Results.}
Ecological Persistence remains at $EP\approx 1.0$ throughout training (Fig.~\ref{fig:chaos_analysis}, bottom), denoting that every computed rollout sustains multiple species above
threshold at every frame, with mean species entropy
$\bar{H}\approx 2.3$\,bits (${\sim}82\%$ of the maximum
$\log_2 7$) (Figure~\ref{fig:chaos_analysis}, top). Effective complexity averages
$\mathcal{C}_{\mathrm{eff}}\approx 0.21\pm 0.05$, above the zero baselines of both ordered and chaotic regimes. Within-rollout trajectories
(Figure~\ref{fig:chaos_analysis}, top) confirm that later runs maintain flat, elevated $\mathcal{C}_{\mathrm{eff}}^{t}$. This shows that PBT-NCA's competitive pressure drives the population toward a stable attractor at the \textit{edge of chaos}, sustaining robust coexistence and structurally rich dynamics.





\section{Discussion}
\label{sec:discussion}

The quest for open-ended complexity remains a defining challenge in artificial life \citep{stanley2019openendedness, taylor2016open}. 
Our results demonstrate that by reframing spatial competition as a novelty-driven, meta-learning process, the PBT-NCA discovers and sustains a diverse manifold of emerging lifelike behaviors mirroring phenomena across computational and biological domains. For instance, in Figure~\ref{fig:pd-nca-ants}, \ref{fig:pbt-nca-emergent} (\textbf{\texttt{meta-iter 310}} and \textbf{\texttt{460}}) and \ref{fig:pd-nca-gliders-shooters}, we observe the spontaneous generation of mobile local structures akin to the ``gun'' and ``gliders'', similarly as in traditional discrete and continuous CA~\citep{gardner70,berto2012cellular,Chan2018LeniaB}.

The system produces complex, macroscopic patterns beyond the scope of localized agents. The traveling spiral waves (Figure \ref{fig:pbt-nca-emergent}, \textbf{\texttt{meta-iter 145}}) resemble the rotating spiral waves in stochastic multi-agent ecosystems \citep{reichenbach2008self} and the nested territorial competition echoes the hypercycles of pre-biotic chemistry \citep{eigen1977hypercycle, cronhjort1995hypercycles}. 
As noted by \citet{crutchfield1995evolution}, the persistence of such structurally stable ``regular domains'' with complex boundaries interacting between them is a prerequisite for emergent computation. This is a promising sign that PBT-NCA may be able to implement computational primitives. Crucially, our framework discovers these complex cycles autonomously, without the pattern-specific handcrafted loss functions required in prior work \citep{pdnca2025}. Furthermore, the spatial segregation into stable macro-structures and volatile frontiers (Figure \ref{fig:pbt-nca-emergent}, \textbf{\texttt{meta-iter 460}}) closely mimics the genetic drift and expanding boundaries of microbial populations \citep{hallatschek2007genetic}. The ejection of ``cell-like'' clusters to colonize new regions (Figure \ref{fig:colonize-collab}) further highlights this biological realism, matching spatial migration observed in nature \citep{baym2016spatiotemporal, nadell2016spatial}. 

Most interestingly, at \textbf{\texttt{meta-iter 230}} of Figure \ref{fig:pbt-nca-emergent}, we observe the emergence of fluid, shape-shifting macro-structures that migrate across the substrate with coordinated, differentiated cell behavior, mirroring motility-induced phase separation~\citep{cates2015motility}. The system does not merely visit these complex states transiently; it sustains cellular differentiation, self-replication, and coordination at the ``edge of chaos''~\citep{langton1990,kauffman1993origins}. 

We hypothesize that the sustained morphological diversity in PBT-NCA is driven by dynamics akin to the Red Queen hypothesis \citep{vanValen1973NewEvolutionaryLaw, trackingRedQueen95, kumar2025drq}. The landscape never settles as individual agents must survive an evolving competitive environment. Our composite novelty score rewards lineages that break away from both historical and contemporary norms. PBT’s exploit-explore mechanism ensures that descendants inherit partially developed ecologies, allowing them to compound complex survival strategies rather than being reinitialized.

\paragraph{Limitations and Future Work.}
While PBT-NCA represents a concrete step toward artificial open-endedness, several limitations remain. The substrate is based on a 2D grid that oversimplifies the thermodynamic constraints driving evolutionary innovation in natural ecosystems \citep{into_the_cool_06, schroedinger1992life, england13}. Additionally, leveraging DINOv2 for visual diversity evaluation may inject anthropocentric, natural-image biases into the system's definition of ``novelty''. 
In the future, we aim to leverage hardware-accelerated frameworks like CAX \citep{faldor2024cax} and hyperscale evolution strategies \citep{sarkar2025evolutionstrategieshyperscale} to support larger grids and more agents \citep{scaling_lenia_23}. This will allow us to study the scaling laws of emergent dynamics in PD-NCA, as well as co-evolving other components such as architectures \citep{Stanley2002EvolvingNN, zoph2017neural, White2023NeuralAS}, environments \citep{wang2019paired, ferreira2022learning, zhang2024omni, bruce2024genie}, and the fundamental update strategies \citep{learning_to_learn_16, chen22L2O}.

\section{Acknowledgements} 

Uljad Berdica is funded by the Rhodes Scholarship and the EPSRC Centre for Doctoral Training in Autonomous Intelligent Machines and Systems. Jakob Foerster is partially funded by the UKRI grant EP/Y028481/1, which was originally selected for funding by the ERC. Jakob Foerster is also supported by the JPMC Research Award and the Amazon Research Award. This project received compute resources from a gracious grant provided by the Isambard-AI National AI Research Resource. Frank Hutter acknowledges the financial support of the Hector Foundation. The authors thank Lukas Seier, Hannah Erlebach, Antonio León Villares, Michael Beukman and Philipp Bordne for their valuable feedback and discussions.


\footnotesize
\bibliographystyle{apalike}
\bibliography{bib/main, bib/extended_refs, bib/bio_references}

\end{document}